\documentclass{article}  
\usepackage{fullpage}

\usepackage[]{graphicx}    
\usepackage{tikz}
\usetikzlibrary{arrows, shapes}
\usepackage{amsfonts}
\usepackage{eqnarray,amsmath}
\usepackage{lipsum}
\newcommand{\ignore}[1]{}  

\usepackage{enumitem}

\usepackage{hyperref}
\usepackage{pdfpages}
\usepackage{xcolor}
\usepackage{subcaption}

\graphicspath{{../Figures/}}

\begin{document}
\title{Graph Spectral Embedding for Parsimonious \\ Transmission of Multivariate Time Series}

\author{Lihan Yao\\
Geometric Data Analytics, Inc. \\
\and
Paul Bendich\\
Department of Mathematics \\
Duke University\\
Geometric Data Analytics, Inc.\\
}

\maketitle

\thispagestyle{plain}
\pagestyle{plain}

\begin{abstract}
We propose a graph spectral representation of time series data that
1) is parsimoniously encoded to user-demanded resolution; 2) is unsupervised and performant in data-constrained scenarios; 
3) captures event and event-transition structure within the time series; and 4) has near-linear computational complexity in both signal length
and ambient dimension. This representation, which we call Laplacian Events Signal Segmentation (LESS), can 
be computed on time series of arbitrary dimension and originating from sensors of arbitrary type. Hence, time series 
originating from sensors of heterogeneous type can be compressed to levels demanded by constrained-communication 
environments, before being fused at a common center.

Temporal dynamics of the data is summarized without explicit partitioning or probabilistic modeling. 
As a proof-of-principle, we apply this technique on high dimensional wavelet coefficients computed 
from the Free Spoken Digit Dataset to generate a memory efficient representation that is interpretable. 
Due to its unsupervised and non-parametric nature, LESS representations remain performant 
in the digit classification task despite the absence of labels and limited data.    
\end{abstract}

\section{Introduction}
\label{sec:Intro}
The historical development of machine learning algorithms on time series data has followed a clear trend from initial simplicity to 
state-driven complexity. For instance, limitations in the Hidden Markov Model (see \cite{zucchini2017hidden} for a survey) for modeling long range dependencies motivated 
the development of more complex but also more expressive neural networks, such as Recurrent Neural Networks \cite{che2018recurrent} or Long Short-Term Memory model \cite{lipton2015learning}.
In this parametric learning framework, 
success in modeling temporal data has been largely dictated by the model's ability to store appropriate latent states in memory, 
and to correctly transition between states according to underlying dynamics of the data. 

Among gradient based models \cite{Jain2015}, 
interpreting these high dimensional latent states, 
as well as relating them to observations, remains a difficult area of research. Moreover, 
modeling dependencies within complex temporal data, such as speech audio \cite{Sargin2007CCA} and financial pricing movements, requires 
a large number of observations, a requirement which often limits their application in real-life scenarios. 

\subsection{The LESS Algorithm}

In this paper, we introduce an interpretable, unsupervised, non-parametric approach to time series segmentation called LESS: 
Laplacian Events Signal Segmentation. 
LESS is motivated by multi-scale geometric ideas, and its core computations are
simple linear algebraic operations and convolution. The algorithm featurizes temporal data in a
  template-matching procedure using
 wavelets. The resulting wavelet coefficient representation is a trajectory in state-space.
 We interpret time steps of this representation as nodes of an underlying graph, whose graph structure is informed by 
 events implicit in the original signal. A partitioning of this graph via its Laplacian embedding 
 results in event segmentation of the signal.  

LESS takes inspiration from the intuitive insight that naturally occurring temporal data have few 
meaningful event types, or motifs, and thus converting signals to event sequences has the potential to drastically reduce transmission 
and storage requirements. We 
aim to derive the simplest unsupervised technique that mirrors existing 
spectral clustering applications in image and graph data domains. 

\subsection{Outline}

After surveying related work in Section \ref{sec:RW}, we give a detailed description of our proposed LESS technique in 
Section \ref{sec:LESS}. Then Section \ref{sec:RaS} gives an empirical analysis of the robustness-to-noise of LESS, as well
as a careful analysis of its computational complexity.

A first application of LESS is shown in Section \ref{sec:Apps}. 
Using the Free Spoken Digits Dataset,
we visualize the temporal dynamics of 
spoken audio with its segmentation into meaningful events, such as strong vs weak enunciations of `rho' in `zero', and show a clear contrast 
  between representation trajectories in wavelet coefficient space belonging to different spoken digits.
Furthermore, we show LESS has superior performance to Dynamic Time Warping \cite{berndt1994using} and to SAX \cite{Lin2007} despite 
summarizing time series observations in far more parsimonious fashion.

Finally, although we do not pursue it formally in this paper, Section \ref{sec:Conc} outlines ways that LESS has strong potential to fit within upstream fusion pipelines of heterogeneous-modality time series.
Hence, we argue that LESS will make a key contribution to unsupervised classification, visualization, and fusion tasks, especially in scenarios
where training data is limited and/or communication between computational nodes is constrained.

\section*{Acknowledgments}
This paper has been cleared for public release, as Case Number 88ABW-2019-4842, 07 Oct 2019.
Both authors were partially supported by the Air Force Research Laboratory under contract AFRL-RIKD FA8750-18-C-0009.
We are grateful to Drs. Peter Zulch and Jeffrey Hudack (AFRL) for motivating discussions, to Christopher J. Tralie for discussions on the scattering transform, to Nathan Borggren and Kenneth Stewart
for pre-processing and discussion of the dataset in Section \ref{sec:RaS}, and to Tessa Johnson for assistance with a clean version of LESS implementation.

\section{Related Work}
\label{sec:RW}

We briefly survey related work, and situate LESS within the context they define.

\subsection*{Connections to wavelet theory} 
Matching pursuit \cite{cai2011orthogonal} projects signal data into its sparse approximation via a dictionary of wavelets. 
By computing the approximation 
error using this dictionary, 
the algorithm greedily selects a new wavelet that maximally reduces this error and adds it to the dictionary. Matching 
pursuit can encode a signal as its sparse approximation using few wavelets. Our algorithm similarly encodes an input signal  
via a wavelet dictionary. However, we further reduce this wavelet representation by computing its 
implicit motifs, summarizing it as a sequence of categorical values.

Similarly, wavelet scattering \cite{mallat2012group} \cite{Bruna2013} also utilizes a wavelet dictionary to derive sparse encodings of signals. In comparison to wavelet 
representations, scatter representations have the additional properties of being signal-translation invariant and 
capturing more complex frequency structure in the signal via its convolution network. We opt for wavelet scattering 
as the signal featurization procedure due to these additional properties. From the perspective of 
wavelet scattering, the LESS representation is in essence 
a temporal segmentation over the scatter coefficients.  

\subsection*{Connections to graph theory} 

There have been numerous works (e.g, \cite{huang2017graph}) that process signals defined on the nodes of a graph.
Tootooni et al \cite{Tootooni2018} propose to monitor process drifts in multivariate time series data using spectral graph-based topological invariants.
Like LESS, the proposed technique 
also considers multidimensional sensor signals and interprets an underlying state graph of the signal. By maintaining 
a window on an incoming stream of multidimensional data, vector time elements within the window form nodes of the graph.
 The authors observe that changes in the Fiedler number, a graph topological invariant computed from the graph's Laplacian matrix, is 
 informative towards process state. The empirical analysis demonstrates effective fault detection in process monitoring applications.    
 LESS differs mainly by applying graph spectral techniques on wavelet-domain representations. In this way, complex
  frequency dynamics within the signal are accounted for in the state graph,
  and an event sequence representation of the signal in its entirety is generated by LESS, instead of a sequence of graph topological statistics
  emitted over the course of a sliding window.  

Lacasa et al \cite{Lacasa2008} propose the visibility algorithm, a process which converts 
time series into a {\it visibility graph}. Visibility graphs preserve the periodicity structure of the time series as graph 
regularity, while stochasticity in the time series are expressed as random graphs by the process. The visibility graph is 
invariant to translation and scaling of the time series, by which affine transformations of the time series lead to
the same visibility graph. In comparison, our approach also generates translation invariant representations due to wavelet scattering.
Periodicity within a time series is recognized as motifs by LESS, and as re-occuring tokens in the resulting event sequence representation.         

\subsection*{Connections to time series modeling} 

Fused LASSO regression for multidimensional time series segmentation \cite{Omranian2015}
combines breakpoint detection with computing breakpoint significance. After determining breakpoints implicit in the time series,
 their technique relies on clustering within each detected segment to estimate breakpoint significance.      
Similar to the proposed technique, LESS also seeks to address the multidimensional time series segmentation problem; 
although we segment according to a set of motifs implicitly determined by the time series, instead of identifying new 
segments for each local change in trend - this allows the desired number of motifs to be preset and easy identification of 
similar events, as they belong to the same motif.        

Sprintz \cite{Blalock2018} is a time series compression technique leveraging a linear forecasting algorithm to be trained online
for a data stream. By combining Delta Coding - a stateless, error-based forecast heuristic common in compression 
literature - with an autoregressive model of the form $x_i = ax_{i-1} + bx_{i-2} + \epsilon_i  $,
the authors are able to predict an upcoming delta as a rescaled version of the most recent delta, leading to a 
more expressive version of Delta Coding.   

This forecasting component of Sprintz named `Fast Integer REgression' (FIRE) is further supported by bit packing and Huffman coding 
to efficiently handle blocks of error values. In comparison, the objective of our paper is to
find structure within time series data by unsupervised learning, then aggressively decrease data representation into a sequence of 
categoric tokens, corresponding to events. While our run-time scales worse by viewing the time series in its entirety, 
LESS representations 
have the flexibility to be of predetermined length, depending on user preference in the granularity of signal segmentation. 
Our representations also convey frequency-temporal structures within the time series.     


SAX \cite{Lin2007} creates a symbolic representation of the time series via Piecewise Aggregate 
Approximation (PAA). By dividing the time series data into equal frames, the PAA representation models the data 
with a linear combination of box basis functions. By binning the PAA coefficients according to the coefficient histogram, 
and setting the global frame size, lengths of SAX symbolic sequences are easily controlled. 
LESS controls the length of its output representation by $k$,
 the set of motifs hypothesized to be within the data. By decomposing the input signal via wavelet scattering,
  LESS featurization presents continuous signal structure in the frequency-temporal domain rather than discrete frames 
  in the temporal domain. Moreover, the resulting wavelet coefficients are segmented by considering its relative location 
  within the time series in wavelet coefficient space, rather than segmentation by binning according 
  to a global coefficient histogram, as is the treatment of PAA coefficients by SAX.           

\section{The LESS Algorithm}
\label{sec:LESS}

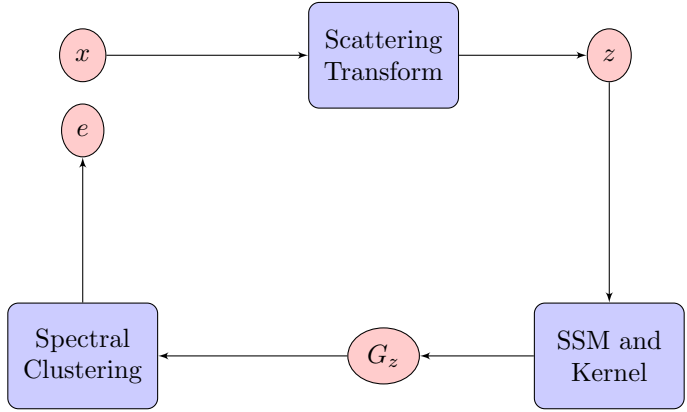
\begin{figure}
\centering
\begin{tikzpicture}[node distance = 4cm, auto]
  \tikzstyle{decision} = [diamond, draw, fill=blue!20, 
    text width=4.5em, text badly centered, node distance=3cm, inner sep=0pt]
  \tikzstyle{block} = [rectangle, draw, fill=blue!20, 
    text width=5em, text centered, rounded corners, minimum height=4em]
  \tikzstyle{line} = [draw, -latex']
  \tikzstyle{cloud} = [draw, ellipse,fill=red!20, node distance=3cm,
    minimum height=2em]
    \node [cloud] (raw) {$x$};
    \node [block, right of=raw] (init) {Scattering Transform};
    \node [cloud, right of=init] (scattered) {$z$};
    \node[block, below of=scattered] (kernel) {SSM and Kernel};
    \node[cloud, left of=kernel] (graph) {$G_z$};
    \node[block, left of=graph] (SC) {Spectral Clustering};  
     \node[cloud, above of=SC] (sequence) {$e$};
    \path [line] (raw) -- (init);
    \path [line] (init) -- (scattered);
    \path [line] (scattered) -- (kernel);
    \path [line] (kernel) -- (graph);
    \path [line] (graph) -- (SC);
     \path [line] (SC) -- (sequence);
\end{tikzpicture}
\caption{A flow diagram for LESS.}
\label{fig:pipeline}
\end{figure}

The high-level steps (Figure \ref{fig:pipeline} shows a flow diagram) in the LESS algorithm are as follows.
First, \emph{wavelet scattering} is applied to a raw time series $x$, resulting in a wavelet representation $z$.
Then a \emph{weighted graph} $G_z$ is computed from the wavelet representation, where the vertices of $G_z$ are 
$z[t], \quad t=1,\ldots,n$. The weighted edges of $G_z$ are computed from the self-similarity matrix of 
$z$ followed by a varying bandwidth kernel. 
Finally, \emph{spectral clustering} is applied to extract an event sequence $e$, the final output of LESS.
This section describes each of these technical steps in details.
Further information on the parameters involved can be found in the Appendix.

\subsection{Wavelet Representation}
\label{sec:WR}


Wavelet scattering \cite{Bruna2013} creates a representation $\Phi x$ of a time series $x$ by composition of wavelet transforms. The 
main utility provided here is reducing uninformative variability in the signal - specifically translation
in the temporal domain 
 and noise components of the data. In the application of event segmentation, it is sufficient for wavelet scattering to capture 
the low frequency structure of $x$. These properties are explained in detail in the rest of this section.
  Throughout this paper, we refer to scatter representation $\Phi x$ as `the wavelet representation', 
  even if the terminology describes a broader family of wavelet-derived objects. 
\begin{enumerate}[leftmargin=0pt, itemindent=20pt,
  labelwidth=15pt, labelsep=5pt, listparindent=0.7cm,
  align=left]
\item $\Phi x$ enables frequency selection \\

The wavelet transform of $x$ is the set of coefficients computed by convolving $x$ with a wavelet dictionary of $J$ wavelets 
\[ \{x * \Psi_\lambda(u)\}_\lambda \]
where each wavelet is indexed by frequency $\lambda$. $\Psi_\lambda$ is convolved 
with the signal to emit coefficients corresponding to frequency $\lambda$. Wavelet scattering re-applies wavelet transform 
on the coefficients of this procedure with the same $q$ wavelets, under the condition that an identical wavelet
cannot be applied to the coefficients more than once. For $p$ applications of wavelet transform, a set of
convolution coefficients are generated, with set size  ${q \choose p}$. We find that a small wavelet dictionary capturing 
low frequency components of $x$ is sufficient, 
leading to fast scatter computations and representations robust to instrumental noise (figure \ref{fig:noise}). 
In practice, $\Phi x$ is most effectively generated by 5 to 20 wavelets of lowest frequency, with $Q=2$.\\

\item $\Phi x$ is invariant to translations of $x$. \\

That is, for a translation $c\in R$ and translated signal $x_c(u) = x(u-c)$,
we have  $\Phi x_c = \Phi x$. As the majority of wavelet techniques are covariant with translation, shifting $x$ in time 
alters the wavelet representation. As we seek to construct an underlying graph $G_z$ capturing the dynamics of $x$, 
translation invariance is an important property for leveraging spectral graph theory techniques. It's relevance emerges 
when considering the instability of Laplacian eigenfunctions $L_{[v1,\ldots,v_e]}$ under a changing graph $G_z$. Despite 
being the same signal, shifts in the temporal domain of $x$ directly replaces vertices of $G_z$, dependent on magnitude of the shift $|c|$
relative to length of the data. 
See below for a detailed discussion.    
\\

\item $\Phi$ linearizes deformations in signal space. \\

For a displacement field $\tau(u)$ (the deformation), $\Phi$ is Lipschitz continuous to deformations if 
there exists $C>0$ such that for all $\tau$ and $x$:
\[ ||\Phi x_\tau - \Phi x|| \leq C ||x|| \sup_u |\nabla \tau(u)| \] 
As an example, for a small deformation in the form of additive noise, $\Phi x - \Phi x_\tau$ is closely approximated by a bounded linear 
operator of $\tau$.
Like translation invariance, $\Phi$'s linearization of deformations in signal space further 
reduces undesirable variability in the construction of $G_z$.   
\end{enumerate}

\begin{figure}
\centering
\includegraphics[width=1\textwidth]{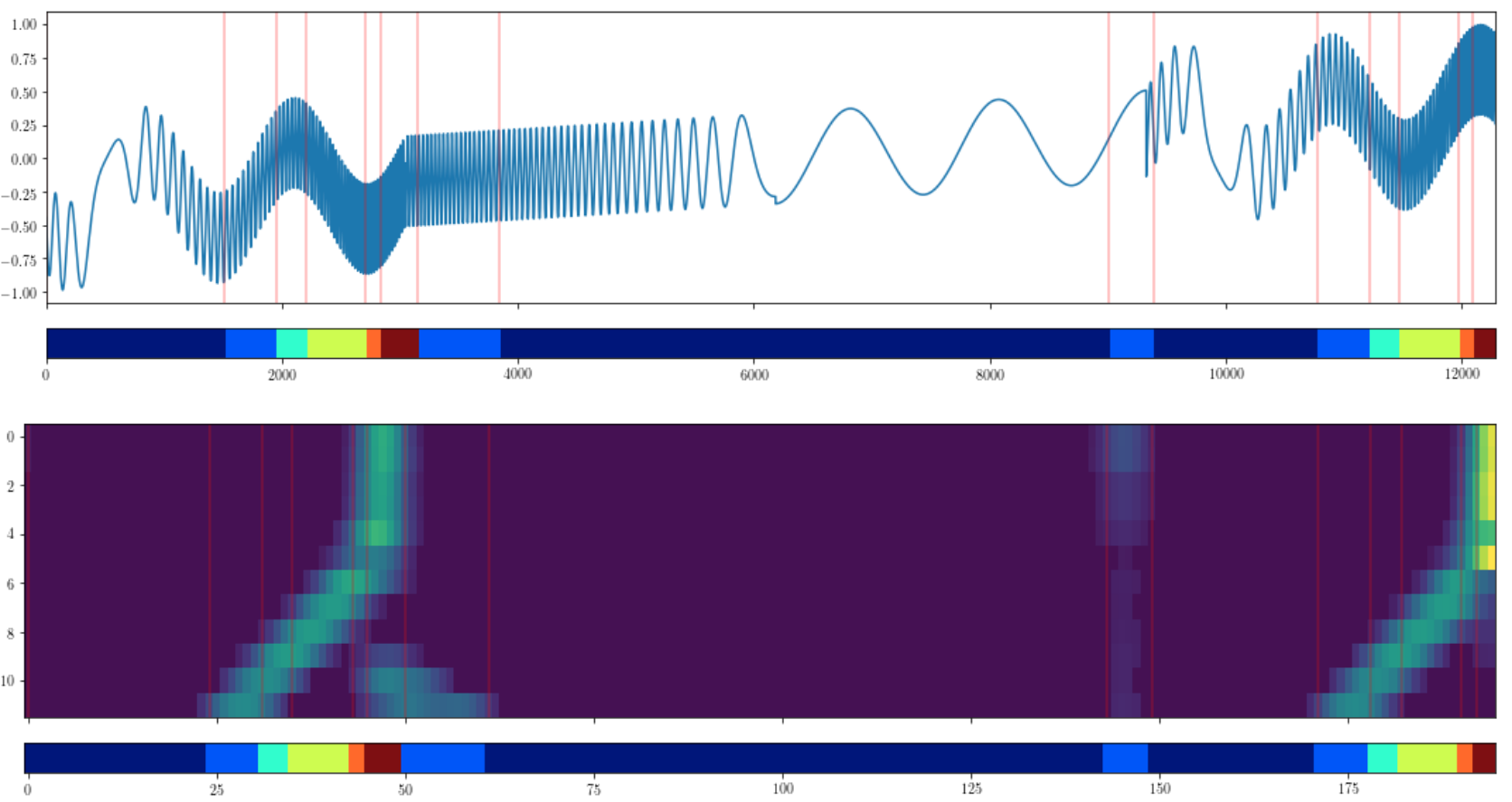}
\caption{{\bf Top} An example signal with it's proposed segmentation $e$ displayed below. The signal begins and concludes 
with sinusoids of increasing oscillation, with an intermediate event. {\bf Bottom} Normalized wavelet representation 
of the above signal. Only coefficients emitted by twelve wavelets of low frequency are considered. Notice the representation is invariant to non-stationarity and local trends, e.g. identical wavelet 
coefficients belonging to yellow events are generated for both a decreasing and an increasing trend in the signal.
In addition, the entire signal follows an upward trend.}
\label{fig:demo}
\end{figure}

Every signal $x$ is mapped by wavelet scattering $\Phi$ to its wavelet representation $\Phi x = z$. In scatter coefficient space 
$Z$, undesirable variability in the space of signals $X$ is removed while the 
frequency-temporal structure of signals is preserved.  

\subsection{Wavelet Trajectory Graph $G_z$}
\label{sec:WG}

Since $z$ may be iterated over by its temporal index $t$, $z$ is also a trajectory in wavelet coefficient space. 
A change in frequency structure of the signal, from here on recognized as an `event', is reflected in $z$ by 
movement within wavelet coefficient space. 
Moreover, re-occuring events return the trajectory to the same region due to their similar frequency characteristics.

\begin{figure}
\centering
  \includegraphics[width=1\textwidth]{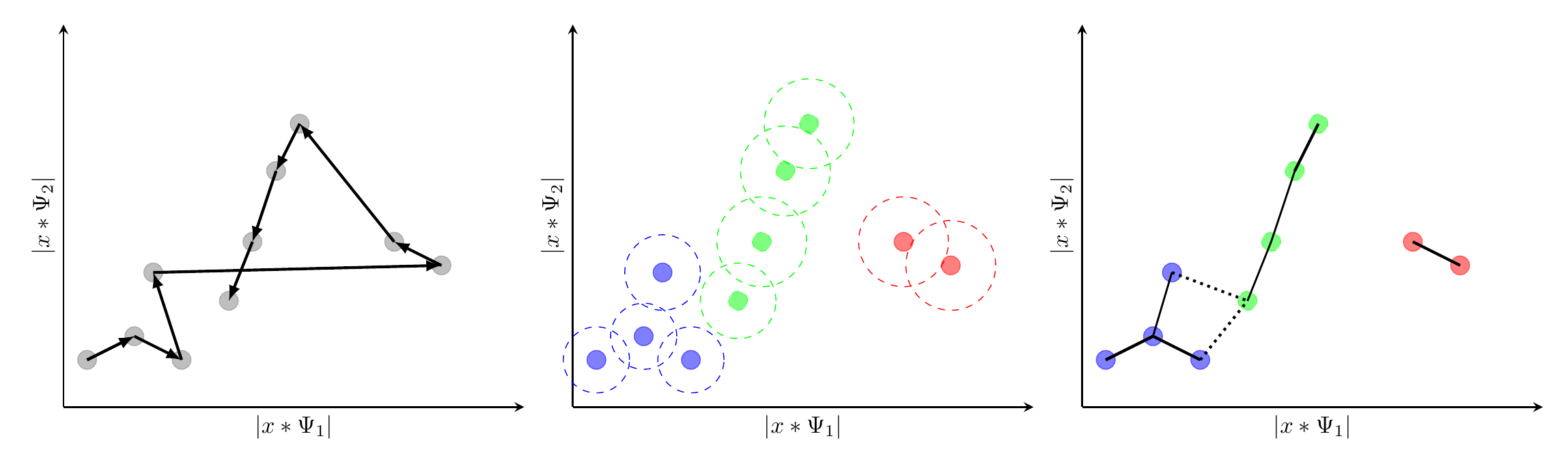}
  \caption{A visualization of spectral clustering in LESS. Figure \ref{fig:wavelet_pca} applies similar analysis on wavelet representations of real data. 
   {\bf Left} The wavelet coefficients of two wavelets is plotted as a trajectory in wavelet coefficient space. {\bf Center}
   Dotted circles denote the range of adaptive kernels. Spectral clustering partitions 
    wavelet coefficient space into non-convex regions according to population density. {\bf Right} Solid lines denote strong edges of $G_z$, dashed lines denote weak edges.}     
   \label{fig:spectral_clustering_illustration}
  \end{figure}

  \begin{figure} 
  \centering
   \includegraphics[width=0.4\textwidth, trim={9cm 4.9cm 9cm 4.7cm},clip]{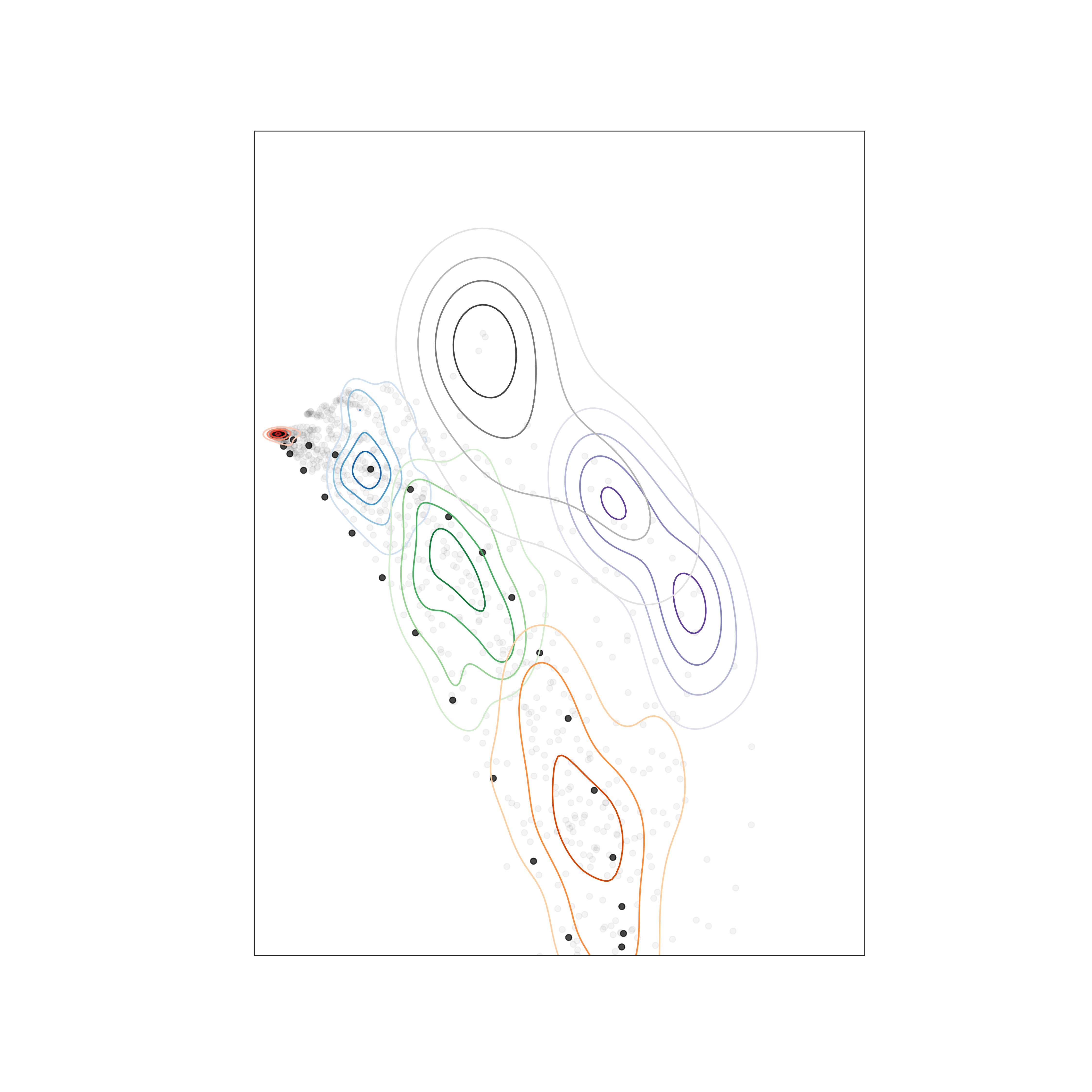}
\includegraphics[width=0.4\textwidth, trim={9cm 4.9cm 9cm 4.7cm},clip]{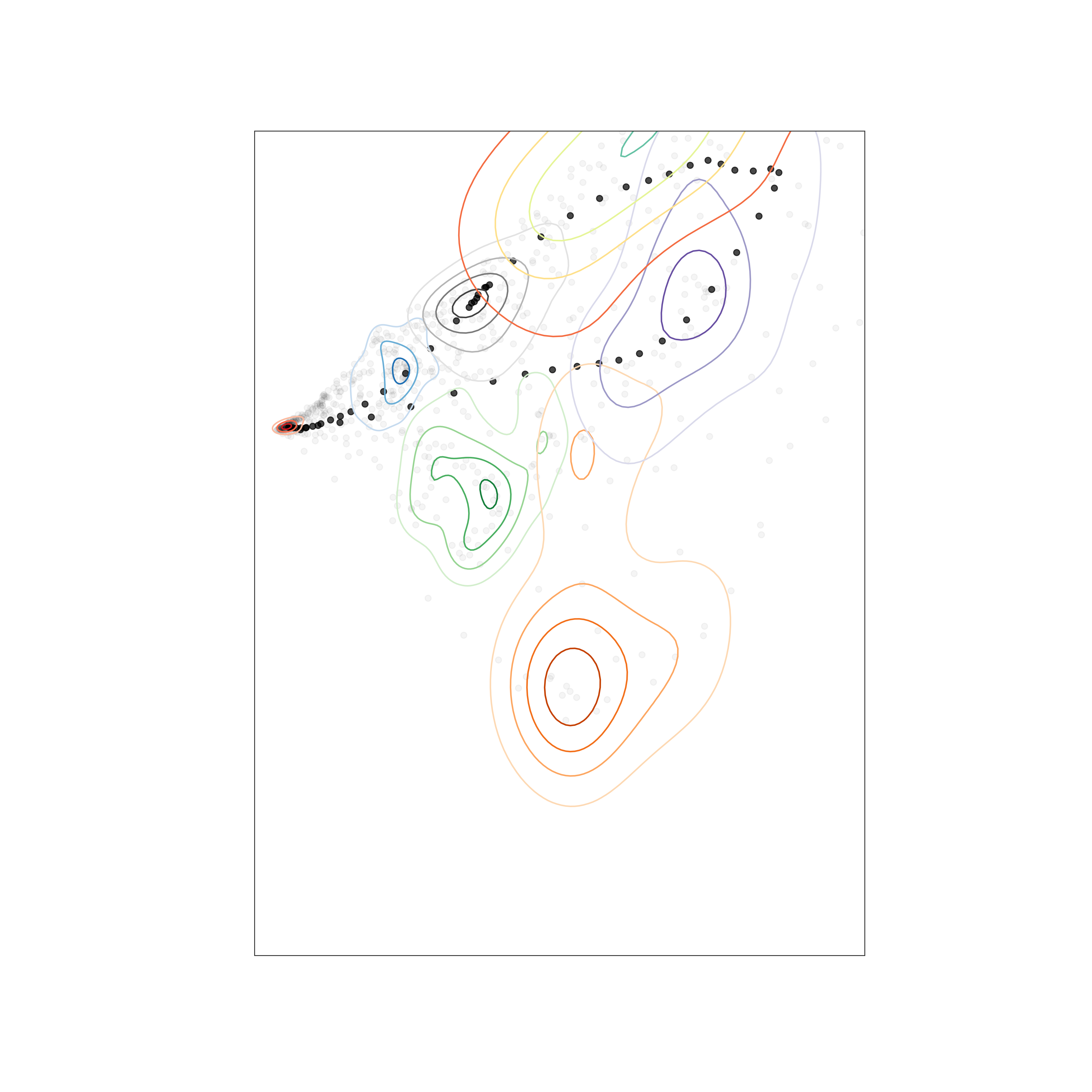}
    \caption{ To interpret the dynamics of wavelet coefficients,
    we project the wavelet coefficient space with time elements in $R^{40}$ to 2 principal components via PCA.
    Colored contours in both figures correspond to the same set of 7 motifs, as mapped by LESS ( red motif captures 
    silence in the audio).
     To compute a common set of motifs among wavelet coefficients of the digit `one' and `two' classes, we concatenate
    the data.
    Differing density between digit classes cause different contour shapes of the same motif.  
    {\bf Left } 20 wavelet representations of the `one' class is plotted in gray, 
    with one example trajectory in black. Notice the bulk of observations do not exhibit black or purple events. {\bf Right }
    20 wavelet representations of the `two' class, with one example in black. This class exhibits an extra event,
     colored in red-green. Wavelet coefficients of this class traverse toward the upper right direction
    in this projected space, differing from `one'. }
    \label{fig:wavelet_pca}
  \end{figure}
  

By segmenting $z$'s traversal to prevalent stationary and transitory patterns throughout 
time, we automate the identification of event motifs. See figure \ref{fig:spectral_clustering_illustration} 
for an illustration, and figure \ref{fig:wavelet_pca} for trajectory plots belonging to real data. 
The role of spectral clustering is identifying prevalent time series motifs. By segmentation of $z$'s dynamics according to 
motif durations, a sequence of events in $x$ is revealed.  

The algorithm constructs an underlying graph $G_z$ of the wavelet representation $z$; the $n$ time elements  
of $z$ form the vertex set of $G_z$ 
\[ V(G_z) = \{z[t]  \hphantom{12}  | \quad t = 1, \ldots, n \} \]
The weighted edges of $G_z$ are determined by the affinity matrix $W$ between time elements. 
The $z[t_a], z[t_b]$ pairwise affinity is

\[W(t_a, t_b)  =  \exp\left( -\frac{ d(z[t_a], z[t_b])^2 }{\sigma_\omega \sigma_{t_a,t_b}} \right) \]  

where $z[t] \in R^p$, $p$ the wavelet dictionary size. $d(\cdot,\cdot)$ is the euclidean distance, 
$\sigma_\omega$ a global scaling parameter. $\sigma_{t_a,t_b}$ is the adaptive kernel size between time elements indexed at $t_a$ and $t_b$.

An adaptive kernel size, in contrast to a global kernel size, mitigates overly or sparsely sampled local neighborhoods. 
Following standard practice \cite{wang2014similarity}, we choose $\sigma_{a,b}$ as adaptive to local neighborhood distances.  
We compute the average distance between $z[t_a]$ and its $C$ nearest neighbors $N_C(z[t_a])$ to approximate the 
local sampling density, and denote this average neighborhood distance by $N_{t_a}$. The adaptive kernel is 
\[ \sigma_{t_a,t_b} = \frac{ N_{t_a} + N_{t_b} + d(z[t_a], z[t_b]) }{3} \]

The interpretation of $z$ as a graph $G_z$ is quite literal. Observe that signal events
 inform the graph structure of $G_z$ - stationary trends in $x$ are easily captured as clique-like substructures in $G_z$, since 
 stationary samples of the signal are all proximal in wavelet coefficient space. 2-state 
 periodic behavior in $x$ may be reflected as a cycle connecting two cliques.

\begin{figure}
\centering
  \includegraphics[width=0.4\textwidth]{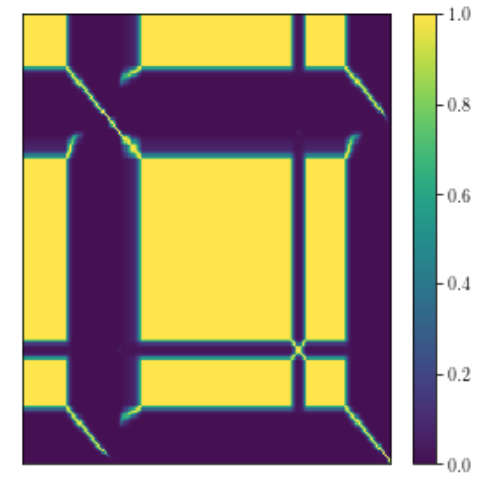}
  \caption{ Affinity matrix $W$ displaying pairwise affinities between time elements of figure \ref{fig:demo}. The matrix center 
  having high affinity is caused by the zero dominated middle interval of the wavelet representation.
   The interruption 
  of this matrix block is caused by a light blue event around sample 9000 in the original signal. Diagonal lines 
  of high affinity in the upper right quadrant (by symmetry of the similarity matrix, also the lower left quadrant) implies 
  a recurrence of wavelet coefficient structure early in the signal with end of signal. }
  \end{figure}

\subsection{Event Identification}
\label{sec:PI}

To exploit this insight on the connectivity and communal graph information, we 
leverage spectral clustering to automate the partitioning of $G_z$, where a subset of time elements $ E \subset V(G_z)$ corresponds to  
an event type in the data. At a high level, this is accomplished by embedding graph vertices by the eigenfunctions of the graph 
Laplacian $L$. Specifically, with adjacency matrix $A$ and degree matrix $D$ of $G_z$, we compute the normalized Laplacian
\[ L = D^{-1/2}AD^{-1/2}  \]
By applying eigendecomposition to $L$, we extract the eigenfunctions of the Laplacian.
$L_{[v_1,\ldots,v_\Gamma]}$, the Laplacian embedding of eigenfunctions corresponding to the smallest Laplacian eigenvalues, 
contains information regarding the stable cuts of $G_z$. See von Luxburg's excellent survey \cite{von2007tutorial}
 for a detailed treatment. 

 $L_{[v_1,\ldots,v_\Gamma]} \in R^{|V(G_z)| \times \Gamma}$ is a `tall' matrix whose rows denote embedding coordinates for individual 
vertices. The eigenfunctions corresponding to larger eigenvalues yield more instable cuts,
and therefore more noisy, vertex partitions.  For most naturally occuring times series, the 
first three eigenfunctions of the graph Laplacian are sufficient.
For the sake of clear notation, by $L_v$ we refer to the low rank embedding
of $L_{[v_1,v_2,v_3]}$.

In $G_z$, each connected component 
 is encoded by a Laplacian eigenvector within eigenspace 0 
 (in other words the geometric multiplicity of eigenvalue 0 matches number of connected components), where an 
 eigenvector is an indicator vector with $1$ if the vertex belongs to that component. For Laplacian eigenvectors
  corresponding to non-zero eigenvalues, they partition connected components within $G_z$.
  The eigenvector of smallest non-zero eigenvalue, the Fiedler vector, lists a partition from the 
 most stable cut of $G_z$ -- a cut partitioning a connected component that acts most like a bottleneck of the 
 component, i.e. in the context of the normalized Laplacian, a cut of maximal flow whose removal leads
  to two components of similar volume.   

Finally, spectral clustering applies $k$-means clustering to $L_{v}$, assigning cluster memberships $\{E_i\}_i$ that partition $ V(G_z)$. 
$\cup_i E_i = V(G_z)$. The number $k$ of cluster centroids, is a LESS parameter dictating the number of motifs to be  
considered while segmenting $x$. For larger $k$, higher time-resolution events may be observed, though overfitting in the form 
of short, spurious events is likely to occur.  

By applying spectral clustering to the wavelet trajectory graph $G_z$, we partition the wavelet representation $z$, and by 
inversely mapping cluster assignments to the original temporal domain of the signal, LESS segments $x$ into a sequence of 
categoric tokens $e$, or \textit{event sequence} of length $l$: 
\[  \text{LESS}(x) = e = \{ e[1], \ldots, e[l]\}  \quad e[i]  \in \{ 1, \ldots, k \} \]      

\section{Properties and Computational Characteristics}
\label{sec:RaS}

\subsection{Moving Average and Noise}
\label{sec:ECG}

This subsection describes two LESS properties: applications to signals with constant moving average and to noisy signals.  

\begin{figure}
\centering
  \includegraphics[width=0.5\textwidth]{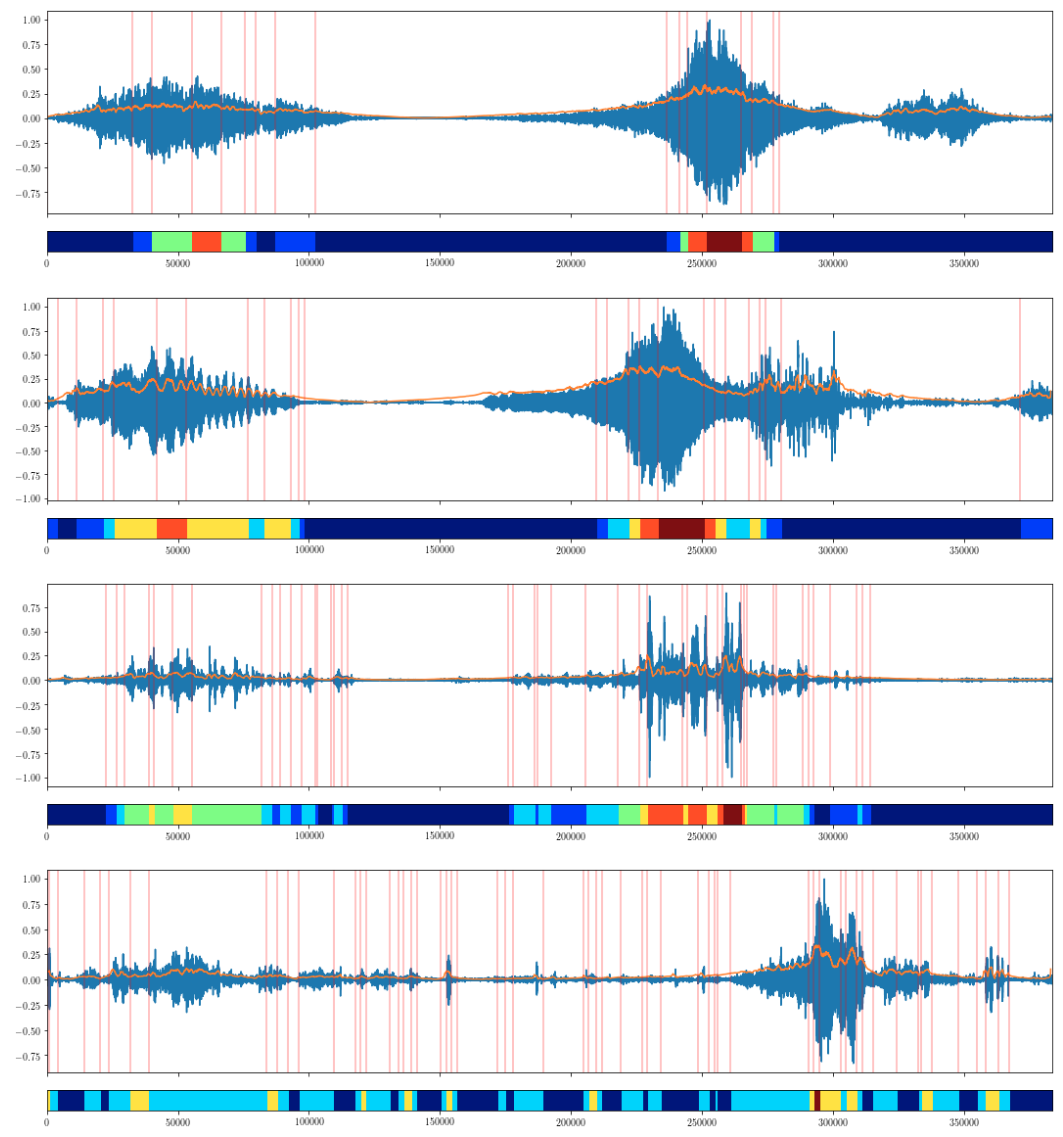}
  \caption{ Four 1-minute audio recordings in the ESCAPE dataset, belonging to single vehicles of different types moving 
  in the same trajectory. THE RMS envelope is shown in orange.}
  \label{fig:escape_annotate}
  \end{figure}

For signals with constant moving average through time, i.e. without trends, LESS event sequences
 tend to reflect local changes in Root Mean Square envelope and energy of the signal. 
To visualize this, 
LESS is applied to the recently-released ESCAPE dataset \cite{ESCAPE}, a collection of live scenes where various vehicles are observed by sensors of different 
modalities. For illustration purposes here, we focus only on the audio modality and on four particular single-vehicle runs with vehicles of distinct types.
In figure \ref{fig:escape_annotate}, four 1-minute audio recordings, belonging to single vehicles of different types,
 are concatenated then annotated by LESS.
In the event sequences below each recording, common colors across observations denote events belonging to the same motif. 

LESS representations remain robust in the presence of significant noise. 
By omitting wavelets corresponding to high frequency indices in  
wavelet scattering, the resulting wavelet representation $z$
 does not account for high frequency structure in the original signal. Thus, while decreasing the number of necessary 
 convolution operations, the algorithm regularizes the representation. 
 
\begin{figure}
  \centering
  \includegraphics[width=0.5\textwidth]{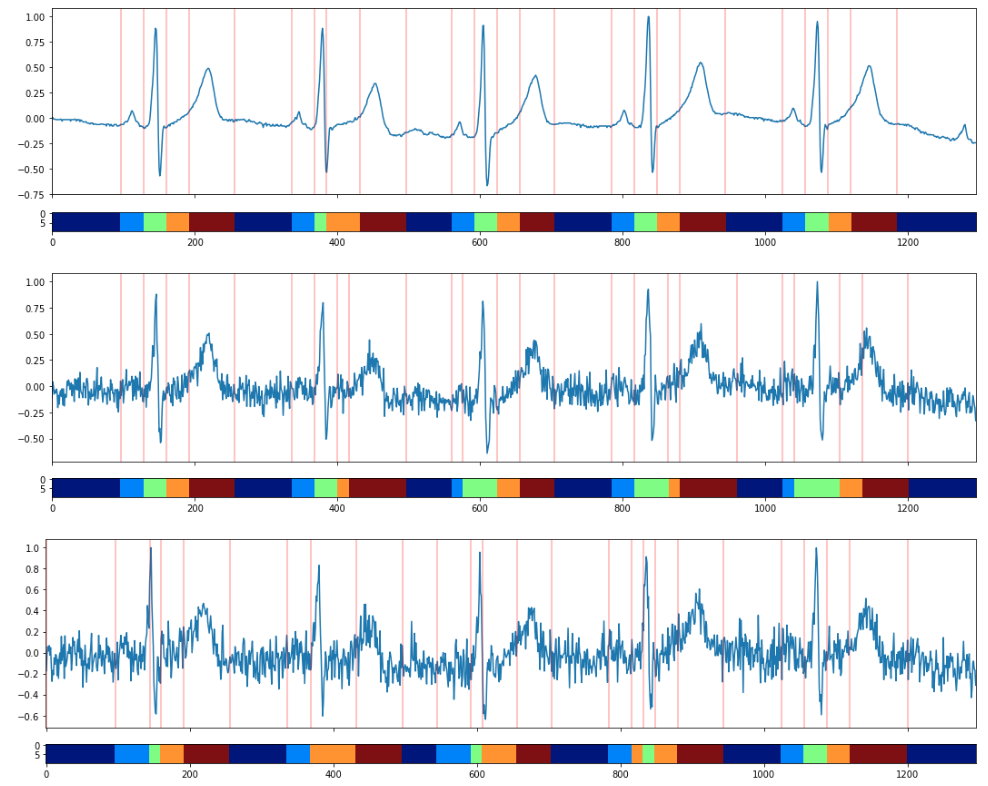}
  \caption{ An observation of the MIT BIH electrocardiogram data with its event sequence. 
  Plots below has added Gaussian noise of variance $\sigma = 0.2$ and $\sigma = 0.5$. The
  cardiac cycle of P wave, QRS complex, ST segment and T wave is consistently identified by the event sequence
 of events light blue, green, orange, and brown respectively. }
  \label{fig:noise}
  \end{figure}

 We illustrate this desirable property via an observation of the MIT BIH electrocardiogram data \cite{MIT-BIH}. 
 As heart muscles contract in a cyclical pattern, the
  cardiac cycle of P wave, QRS complex, ST segment and T wave is consistently identified by the event sequence
 of events light blue, green, orange, and brown respectively.   
 In addition to the ECG signal, Gaussian noise of variance $\sigma = 0.2$
 and $\sigma = 0.5$ was added. As seen in Figure \ref{fig:noise}, LESS has identified all noticeable patterns within
  the original signal accurately; moreover, it continues to output similar annotations under increasing noise.   
  
\subsection{Computational Complexity and Scaling with Ambient Dimension}
\label{sec:CI}


We analyze the computational load of LESS in this subsection. Note that LESS first applies wavelet scattering and then spectral clustering. 
In case of a multivariate signal, LESS computations scale linearly with size of the ambient dimension. Altogether, we claim that LESS has the following computational complexity 

\[O\left(D t \log t \right) + O(n^3), \]

where $D$ is the number of ambient dimensions, $t$ the length of the time series, 
and $n$ the number of time steps in wavelet representation. We argue below that $n \ll t$ in general practice, and thus the cubic term in the complexity analysis should not be that daunting.   

To see this claim, we note (see \cite{mallat2012group})
that scattering computation on $1$-dimensional signals can be done in $O\left(\frac{1}{3}^m t \log t \right),$
with a signal of length $t$ and a scatter network of depth $m$.
In practice, scatter networks of depth $m=2$ are optimal. Given a multivariate signal
in $R^D$, we may naively apply the scatter computation $D$ times to generate coefficients 
for each dimension of the input, showing that the wavelet scattering of multivariate time series has complexity 
$O\left(D t \log t \right).$
In practice, we note that modern numeric packages (e.g, kymatio \cite{kymatio})
for wavelet techniques leverage Graphic Processing Units (GPUs) to accelerate convolution operations.   

The remainder of the computation concerns neither the number of ambient dimensions $D$ nor the length of input data $t$. 
During wavelet scattering, the extraction of wavelet coefficients is preceded by convolution with a low pass filter
 $\phi_{2^J}$, where the original length is sampled at intervals $2^J$, $J$ a scaling parameter set prior. Given a length 
 $t$ signal, its wavelet representation has $n$ time steps: with increasing $J$, $n$ decreases exponentially relative to $t$. 
 For most reasonable 
 values of $J$, we have $ n \ll t$.        

Given a $D$ dimensional time series, there are $D$ applications of wavelet scattering, each application with the same wavelet dictionary 
of size $p$. What results 
is a dimension-wise concatenated 
wavelet representation of $Dp \times n$ coefficients. To compute the affinity matrix for spectral graph analysis,
the $n$ vectors in $R^{Dp}$ are subtracted pairwise. Using vectorized subtraction, the distance matrix is computed 
in $O(n^2)$.  
After transferring high dimensional coefficient information into connectivity relationships in the underlying graph $G_z$,
 the spectral clustering portion of the algorithm becomes agnostic to the ambient dimension of the input time series.    

Lastly, spectral clustering's computational bottleneck arises in acquiring a low rank approximation to the affinity matrix.
Using the Nystrom method \cite{li2011time}, there is an inherent two step orthogonalization procedure,
where $O(n^3)$ operations is incurred. In total, LESS operates with computational complexity $O\left(D t \log t \right) + O(n^3),$ as claimed.

\section{Application}
\label{sec:Apps}

This section outlines a first proof-of-principle application for LESS.
We focus on distinguishing spoken digits from one another, using audio recordings, in an entirely unsupervised manner.
That is, we use LESS to derive a distance between spoken audio observations, and then observe that clusters formed by 
this distance tend to correspond to digit type.
The performance is especially encouraging when compared with two other common
 methods for deriving distances between time series data, Dynamic Time Warping and SAX.

The \href{https://github.com/Jakobovski/free-spoken-digit-dataset}{Free Spoken Digits Dataset}(FSDD) consists of spoken digit recordings 
in wav files at 8kHz. FSDD contains 2,000 recordings from 4 speakers, each speaker saying digits 0 to 9 ten times per digit. 

\begin{figure}
\centering
  \includegraphics[width=0.5\textwidth]{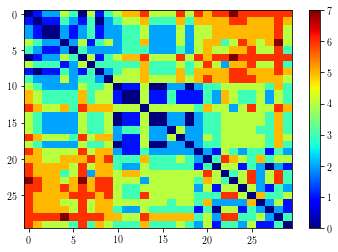}
  \caption{ Levenshtein distances between event sequences. Contiguous slices of 10 rows/columns
  correspond to representations belonging to classes `zero', `one', and
  `three' respectively. Lengths of event sequences range from four events (tokens) to eight.
  }
  \label{fig:013}
  \end{figure}

Representations derived from the proposed technique remain informative towards classification tasks. To illustrate this, we compute 
within-class and between-class distances between event sequences. As example, given parameters $\theta$ and 
$k=3$ possible event types, $\{\alpha, \beta, \gamma \}$, LESS maps signal $x$ to an event sequence
\[ { 
[\alpha, \beta, \alpha, \gamma, \ldots, \beta]} \]
Such an event sequence contains categoric tokens corresponding 
to changes in the frequency structure of $x$, and consecutive tokens of the same motif are replaced with one in their place. 
The Levenshtein distance \cite{levenshtein1966binary} is a standard method to compare any two token strings.
Figure \ref{fig:013} shows the Levenshtein distance matrix computed between LESS representations of 30 short audio strings, 10 each of digit
types '0', '1', and '3.'
The rows of the matrix are ordered by digit type.
For our choice of hyperparameters,
the event sequence lengths range from four events (or tokens) to eight, ensuring transmission parsimony. Within each class,
our technique generated event sequences that are, in expectation, 2 edits away. As seen in the block diagonal structure of the matrix,
  between class distances are higher.           

We compare to two other methods of computing distances between signal snippets, Dynamic Time Warping (DTW, \cite{berndt1994using}) and SAX \cite{Lin2007}. In brief, DTW produces an optimal match between the time indices of two time series observations that is robust to 
 non-linear temporal distortions, in addition to temporal translation. Given a pair of signals $(x_i,x_j)$, the \textit{DTW distance} refers to the minimal cost
  required to align time indices between $x_i$ and $x_j$, or equivalently the matching cost of their optimal match.
  Note that DTW can be computed on pairs of time series in \emph{any} metric space; 
  here we use it both in signal space $X$ and in wavelet coefficient space $Z$.
 SAX, on the other hand, creates a symbolic representation of the time series via Piecewise Aggregate 
Approximation (PAA). The output of SAX is very similar to that of LESS in form, namely a sequence of categorical tokens. Two such token sequences can then be compared using Levenshtein distance as above. 

\begin{figure}  
\begin{subfigure}{0.3\textwidth}
\includegraphics[width=\linewidth]{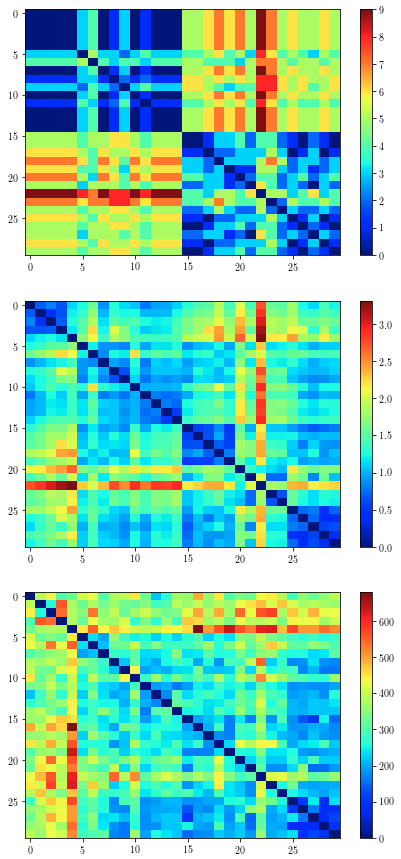}
\end{subfigure}
\hfill 
\begin{subfigure}{0.3\textwidth}
\includegraphics[width=\linewidth]{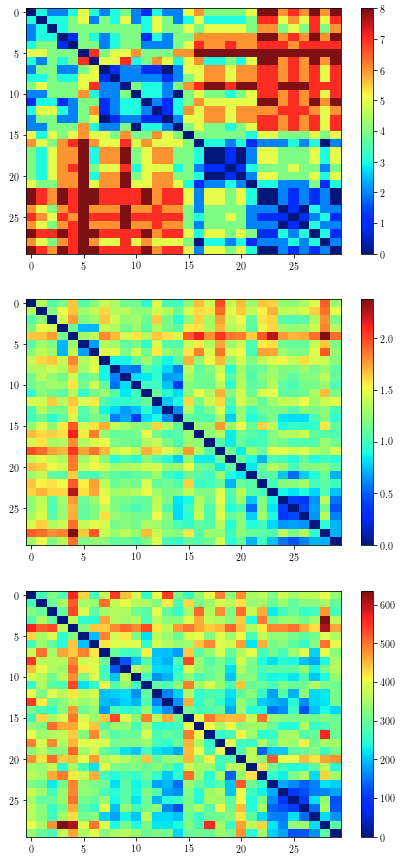}
\end{subfigure}
\hfill 
\begin{subfigure}{0.3\textwidth}
\includegraphics[width=\linewidth]{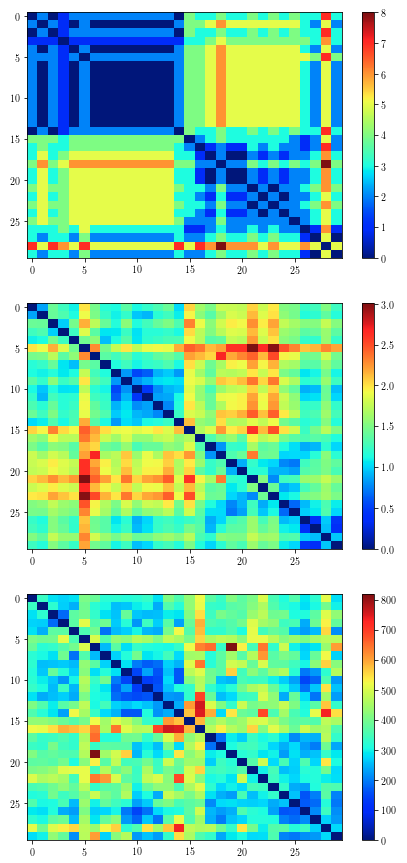}
\end{subfigure}
\caption{ Each column corresponds to an experiment. Left column lists distances between spoken digits
 `one' (row/col indices 0 to 14) and `six' (row/col indices 15 to 29). Center column lists distances between `four' and `seven'. 
Right column lists distances between `two' and `five'.  
{\bf Top} Levenshtein distance between LESS event sequences 
{\bf Center} Distances approximated by Fast DTW when it is applied to wavelet representation time steps $z[t], \quad t = 1,\ldots, n$  
{\bf Bottom} Signal distances approximated by Fast DTW. 
Notice in experiment 1, the observation at index 22 was considered far from all others in every distance. }
\label{fig:3X3_dtw}
\end{figure}

\begin{figure}  
  \begin{subfigure}{0.3\textwidth}
  \includegraphics[width=\linewidth]{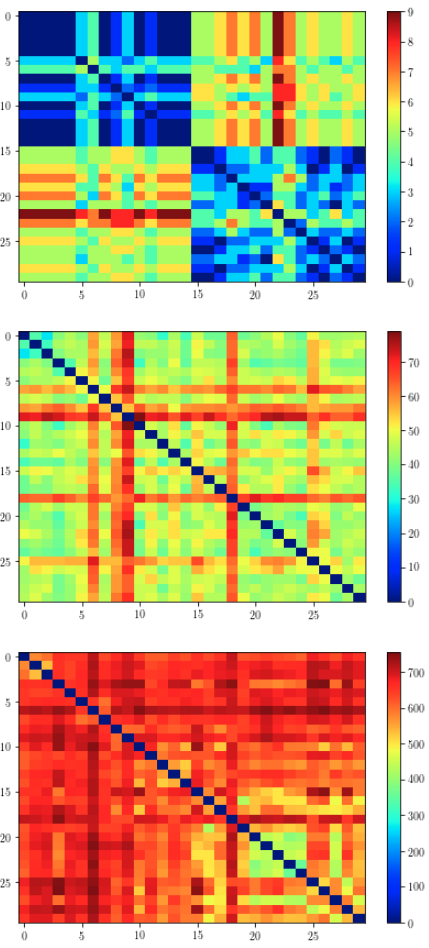}
  \end{subfigure}
  \hfill 
  \begin{subfigure}{0.3\textwidth}
  \includegraphics[width=\linewidth]{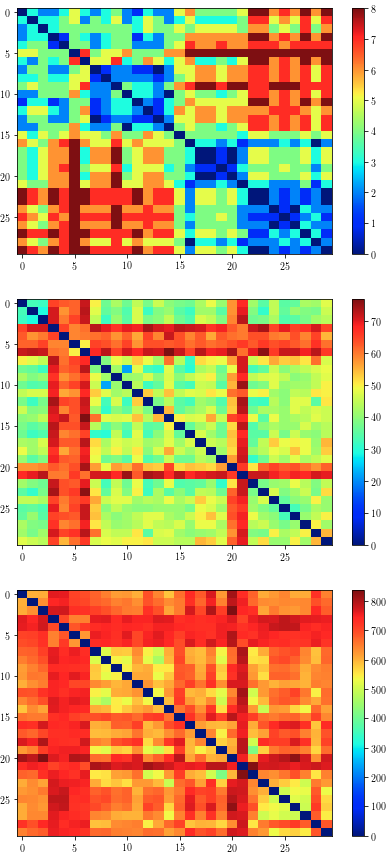}
  \end{subfigure}
  \hfill 
  \begin{subfigure}{0.3\textwidth}
  \includegraphics[width=\linewidth]{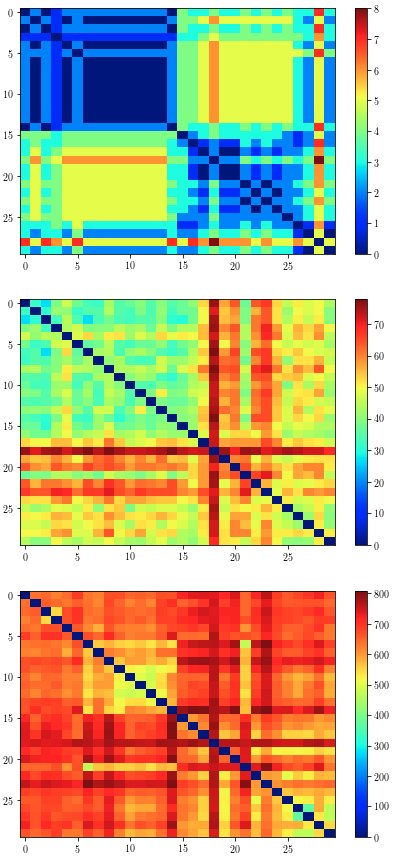}
  \end{subfigure}
  \caption{ Each column corresponds to an experiment. Left column lists distances between spoken digits
   `one' (row/col indices 0 to 14) and `six' (row/col indices 15 to 29). Center column lists distances between `four' and `seven'. 
  Right column lists distances between `two' and `five'.  
  {\bf Top} Levenshtein distance between LESS event sequences, lengths varying from $4$ to $10$. 
  {\bf Center} SAX distance for alphabet size $10$ and representation length $100$.  
  {\bf Bottom} SAX distance for alphabet size $10$ and representation length $1000$.}
  \label{fig:3X3_sax}
  \end{figure}

 Figures \ref{fig:3X3_dtw} and \ref{fig:3X3_sax} shows the results of the comparison experiments.
Each column of each figure corresponds to a single experiment, where the task is to distinguish between $30$ spoken instances of two digit types.
 The top row in both figures uses the LESS-Levenshtein distance.
 The middle and bottom rows of Figure \ref{fig:3X3_dtw} use DTW computed on the scattering space time series and the raw time series, respectively.
Qualitatively, the advantage of the LESS technique over the others at this task is clear.
Furthermore, there are two other practical advantages: 1) in wallclock time,
 the approximal DTW algorithm, Fast DTW \cite{salvador2007toward}, took an order of magnitude of time longer 
 than LESS while making $ \binom{30}{2}$ comparisons; 2)   
 the advantage of LESS over DTW in a communications-constrained environment should be clear: before DTW is computed on two signals arising from different sources, the entire signal must be transmitted, while LESS requires transmission of only the much shorter event sequence.
 
 Figure \ref{fig:3X3_sax} shows the same experiments in a comparison with SAX. A SAX parameter 
 alphabet size is the number of token types available for labeling fixed width windows. 
 We set alphabet size to 10 following 
 the recommendation of SAX authors. We set representation lengths to be 100 (middle row)
  and 1000 (bottom row). For representation lengths of 1000, 
  within-class structure begins to emerge in distance matrices. Again, the practical advantages of LESS 
  in terms of computational speed and transmission in communications-constrained environments hold.     
 
\section{Discussion}
\label{sec:Conc}

This paper presented LESS: Laplacian Events Signal Segmentation, a graph spectral representation of arbitrary-dimensional time series data.
Despite its unsupervised nature, LESS is shown to be highly performant at a digit classification task, and especially so when judged in
data and/or communication constrained environments.
Further work will proceed along several fronts:

\subsection{Memory Issues}

Despite the nice computational complexity analysis above, the current implementation of LESS presents
serious memory management issues on conventional hardware.
An application of LESS requires the concatenation of multiple signals under consideration into one array, due to 
spectral clustering's memory-less nature.
This is impractical for large batches: 1) the eigendecomposition term in the complexity analysis $O(n^3)$ dominates 
the computation and 2) $G_z$ may not be stored for the entire dataset $X$. 
Hence, on a desktop computer with conventional hardware,
LESS may process 5-10 minutes of audio sampled at 48,000 Hz within 30 minutes. 
 
Proximal and batch versions of spectral
clustering \cite{han2017mini} \cite{yan2009fast} offer solutions to alleviate this, both for maintaining adequate 
memory to store the graph $G_z$ and for tackling matrix computations on large batches of data. We will pursue these improvements
in the near future.

\subsection{Towards LESS as a Fusion Technique}

The experiments above only display the benefits of LESS as a compression technique for tokenizing high-dimensional time series before transmission down a stingy channel.
Beyond this paper, we are developing methods to use LESS within upstream fusion pipelines.
The most immediate approach is to note that LESS involves the computation of a weighted graph based of a distance matrix (see the $G_z$ block in Figure \ref{fig:pipeline}). 
When faced with $n$ distinct time series ${\bf x}_1, \ldots, {\bf x}_n$ of arbitrary dimensionality, one could simply run LESS up to this block, producing
$n$ weighted graphs $G_{z_1}, \ldots G_{z_n}$.
Any number of distance-graph-based upstream fusion techniques (e.g, similarity network fusion \cite{wang2014similarity}, \cite{Tralie2019SNF} or joint manifold learning \cite{Davenport2010HighDD}, \cite{ShenJML2018}) could then be used to produce a fused weighed graph $G_f$. 
The final step of LESS could then be applied to produce the fused event sequence.
Experiments must be done to show that the resulting event sequence is indeed more informative, at tasks such as the ones outlined above, than stovepiped event sequences.

In a communication-constrained environment, the transmission of entire weighted graphs $G_{z_i}$ might be too expensive.
Hence further work needs to be done on this front, either by: 1) pursuing sparse representations of the weighted graph, or; 2) creating an event-sequence-level fusion algorithm.

\newpage

\bibliographystyle{plain}
\bibliography{GDAreferences}

\newpage
\section{Appendix}
\label{sec:Imp}

This appendix gives more details on the parameters involved in LESS.
All parameters are inherited from wavelet scattering and spectral clustering.

This paper implements wavelet scattering from the Kymatio package \cite{kymatio}, and the parameter notations 
are adopted from Kymatio documentation. Scaling parameter $J \in Z$ dilates morlet wavelets by a factor of $2^J$. As a 
wavelet's sinusoid component dilates, convolution with the signal leads to decorrelated scales. This reveals 
the frequency structure of the signal at a higher resolution, while decreasing temporal resolution. In practice $J$ in the 
range $[6, 10]$ is suitable for most signal data, while short signals require $J<6$ due to the lack of adequate 
temporal support for wavelets. $Q$ is the number of first-order wavelets per octave. For most applications, incrementally exploring $q$ 
by multiples of $16$ seems efficient. In practice, selecting the $16$ wavelets indexed by lowest frequencies are sufficient.

$\sigma_\omega$, the spectral clustering parameter found in affinity matrix computations, controls the notion of
 global proximity. 
 For increasing $\sigma_\omega$, kernel radii surrounding points expand, resulting in larger edges in $G_z$. For 
 computing kernel sizes of a normalized pairwise-distance matrix, $\sigma_\omega \in [0.2, 0.7]$ is optimal. 
 
 By examining the 
 Laplacian eigenvectors $L_{[v_1,\ldots,v_\Gamma]}$, $G_z$ vertices are encoded into the Laplacian embedding $R^\Gamma$. For all 
 LESS experiments shown, only $\Gamma = 3$ has been used; but membership information contained in eigenvectors 
 of larger eigenvalues may prove beneficial. 
 
 Lastly $k$, the number of motifs, 
 is the number of clusters in $k$-means clustering applied onto the embedding $L_{[v_1,\ldots,v_\Gamma]}$. In simple signal data 
 that rarely exhibit novel events, such as FSDD, $k=7$ is sufficient. On the other hand, 
 to transform entire audio scenes, larger $k$ ($\geq 12$) is required.     

The event sequence $e$ may be interpreted as annotation of the $n$ length wavelet representation $z$. While running
various classification tasks, we find $e$ remains performant after discarding consecutive tokens of the same motif,
 and only note when the event type has changed. 

\newpage

%
%
%

\end{document}